\documentclass[onecolumn]{IEEEtran}
\IEEEoverridecommandlockouts
\usepackage{cite}
\usepackage{amsmath,amssymb,amsfonts}
\usepackage{algorithmic}
\usepackage{graphicx}
\usepackage{textcomp}
\usepackage{xcolor}
\usepackage{enumitem}
\usepackage{float}
\usepackage{makecell}
\usepackage{multirow}
\usepackage{booktabs} 
\usepackage{booktabs}
\usepackage{placeins}
\usepackage[font=normalsize]{caption} 
\def\BibTeX{{\rm B\kern-.05em{\sc i\kern-.025em b}\kern-.08em
    T\kern-.1667em\lower.7ex\hbox{E}\kern-.125emX}}
\begin{document}

\title{Condensation-Concatenation Framework for Dynamic Graph Continual Learning\\}

\author{\IEEEauthorblockN{1\textsuperscript{st} Tingxu Yan}\\
	\IEEEauthorblockA{\textit{College of Computer and Information Science} \\
		\textit{Southwest University}\\
		ChongQing, China \\
		929549561@qq.com}\\
\and
\IEEEauthorblockN{2\textsuperscript{nd} Ye Yuan*}\\
\IEEEauthorblockA{\textit{College of Computer and Information Science} \\
	\textit{Southwest University}\\
	ChongQing, China \\
	*yuanyekl@swu.edu.cn}
}

\maketitle

\begin{abstract}
Dynamic graphs are prevalent in real-world scenarios, where continuous structural changes induce catastrophic forgetting in graph neural networks (GNNs). While continual learning has been extended to dynamic graphs, existing methods overlook the effects of topological changes on existing nodes. To address it, we propose a novel framework for continual learning on dynamic graphs, named Condensation-Concatenation-based Continual Learning (CCC). Specifically, CCC first condenses historical graph snapshots into compact semantic representations while aiming to preserve the original label distribution and topological properties. Then it concatenates these historical embeddings with current graph representations selectively. Moreover, we refine the forgetting measure (FM) to better adapt to dynamic graph scenarios by quantifying the predictive performance degradation of existing nodes caused by structural updates. CCC demonstrates superior performance over state-of-the-art baselines across four real-world datasets in extensive experiments.
\end{abstract}


\section{Introduction}
Graphs serve as fundamental structures for modeling relational data in domains like social networks. Graph Neural Networks (GNNs) have become the standard framework for graph representation learning with significant success. However, real-world graphs are dynamic, continually evolving through the addition and deletion of nodes and edges. This temporal dimension challenges conventional GNNs designed for static graphs, leading to catastrophic forgetting where learning new patterns overwrites previously acquired knowledge.

Although continual learning research has expanded to dynamic graphs, evaluation frameworks lack graph-specific adaptations. Metrics like forgetting rate, borrowed from static domains, fail to capture structural cascading effects. Existing methods employ broad preservation strategies while overlooking that topological changes influence representations of numerous existing nodes.

To tackle these limitations, we propose the Condensation-Concatenation Framework for Dynamic Graph Continual Learning (CCC) framework. CCC compresses historical graph snapshots into compact semantic representations and integrates them with current embeddings through concatenation. This approach aims to capture structural change propagation while maintaining representation stability.

Our contributions are:
\begin{itemize}[leftmargin=*]
\item Identification of limitations in existing continual learning evaluation metrics for dynamic graphs.

\item Proposal of the CCC framework combining graph condensation with feature concatenation.

\begin{figure*}[h]  
	\centering  
	\includegraphics[scale=0.15]{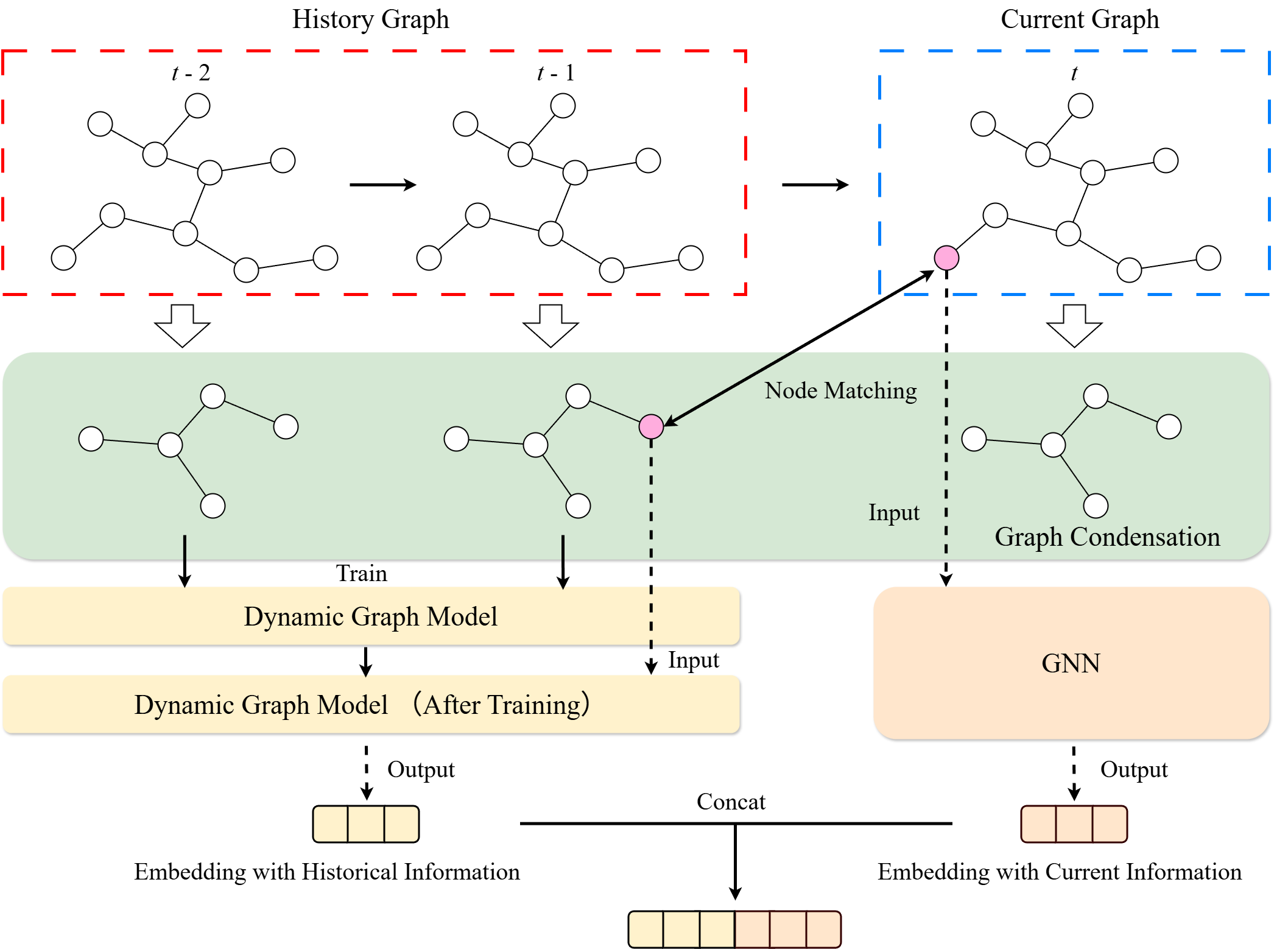}  
	\caption{An overview of CCC. The historical graph sequence is first condensed to capture historical information. Subsequently, by detecting the k-hop structural change regions triggered by node and edge additions/deletions, the embeddings extracted from the condensed historical graph are selectively concatenated with the current embeddings of the affected nodes.}  
	\label{fig:example}  
\end{figure*}

\item Comprehensive experimental validation demonstrating superior performance in balancing knowledge retention with new information integration.
\end{itemize}
\section{Related Work}
\noindent\textbf{Graph Neural Networks.} Graph Neural Networks (GNNs)\cite{b1} aim to apply deep learning to graph-structured data. The core of their approach is to aggregate information from neighboring nodes for learning node representations. Graph Convolutional Network (GCN)\cite{b2} established a spectral graph convolution framework through first-order neighborhood approximation. GraphSAGE\cite{b3} proposed a framework based on sampling and aggregation instead of using all neighboring nodes.In addition, numerous other GNN studies\cite{b4,b5,b6,b7,b8,b9,b10,b11,b12,b13,b14,b15,b16,b17,b18,b19,b20,b21,b22,b23,b24,b25,b26,b27,b28,b29,b30,b31,b32,b33,b34,b35,b36,b37,b38,b39,b40,b41,b42,b43,b44,b45,b46,b47,b48,b49,b50,b51,b52,b53,b54,b55,b56,b57,b58,b59,b60,b61,b62,b63,b64,b65,b66,b67,b68,b69,b70,b71,b72,b73,b74,b75,b76,b77,b78,b79,b80} have also made significant contributions and proven valuable in various graph learning tasks.

\noindent\textbf{Continual Learning.} Continual learning allows models to learn from sequentially arriving data while avoiding catastrophic forgetting of previously acquired knowledge. Existing approaches can be categorized into three groups:

Regularization-based methods protect learned knowledge by incorporating constraints into the loss function. TWP\cite{b81} applies regularization based on parameter sensitivity to topological structures. DyGRAIN identifies affected nodes from a receptive field perspective for selective updates. 

Memory replay-based methods consolidate knowledge by storing and replaying historical data. ER-GNN\cite{b82} adopts multiple strategies to select replay nodes. SSM uses sparsified subgraphs as memory units to retain topological information. DSLR\cite{b83} selects replay nodes based on coverage and trains a link prediction module. PUMA condenses original graphs into memory graphs and retrains on them.

Parameter isolation-based methods allocate dedicated parameters for new tasks. PI-GNN identifies stable parameters through knowledge rectification and expands new parameters in an isolation stage to learn changing patterns.

Our method can be cast as a memory replay-based method. It uses embeddings generated from condensed historical graphs as replay, providing historical information for nodes with altered neighborhoods to mitigate catastrophic forgetting.

\section{Method}
This section presents the details of the CCC method, as illustrated in Figure 1.

\noindent\textbf{Problem Formulation.} The dynamic graph system is denoted as $\mathcal{G}=\{ \mathcal{G}^{(1)}, \mathcal{G}^{(2)},\dots, \mathcal{G}^{(T)} \}$, where each time step \textit{t} corresponds to a graph structure $\mathcal{G}^{(t)} = (\mathcal{V}^{(t)}, \mathcal{E}^{(t)})$. Here, $\mathcal{V}^{(t)} = \{ v_1, v_2, \dots, v_n \}$ represents the set of nodes involved in interactions at time \textit{t}, and $\mathcal{E}^{(t)} = \{ e_{ij} \mid v_i, v_j \in \mathcal{V}^{(t)} \}$ records the connection relationships between nodes at time \textit{t}.

The graph topology evolves over time, with the elements and size of $\mathcal{V}^{(t)}$
 and $\mathcal{E}^{(t)}$ potentially changing. Accordingly, the system includes a sequence of adjacency matrices $\mathcal{A} = \{ \mathbf{A}^{(1)}, \mathbf{A}^{(2)}, \dots, \mathbf{A}^{(t)} \}$ , where each $\mathbf{A}^{(t)} \in \{0,1\}^{N^{(t)} \times N^{(t)}}$
 is defined such that $\mathbf{A}^{(t)}_{ij} = 1$ if and only if $e_{ij} \in \mathcal{E}^{(t)}$.
 
 The node feature sequence is denoted as $\mathcal{X} = \{ \mathbf{X}^{(1)},\mathbf{X}^{(2)}, \dots, \mathbf{X}^{(t)} \}$, where $\mathbf{X}^{(t)} \in \mathbb{R}^{N^{(t)} \times d}$ contains the feature vectors of all nodes, with each node $v_i$ having a feature vector $\mathbf{x}_i^{(t)} \in \mathbb{R}^{1 \times d}$. Here, $N^{(t)}$ refers to the number of nodes at time step \textit{t}, and \textit{d} is the dimensionality of the feature vector for each node.
 
 For the graph neural network model, the input data at each time step is represented by the adjacency matrix and node features, i.e., $\mathcal{D}^{(t)} = (\mathbf{A}^{(t)}, \mathbf{X}^{(t)})$.
 \vspace{5pt}

 \noindent\textbf{Graph Condensation.} The condensing process first generates the node set of the condensed graph based on the distribution ratio of node labels. Given the original graph $\mathcal{G} = (\mathcal{V}, \mathcal{E})$, the node set of the condensed graph \( \mathcal{V}' \) is generated proportionally according to the frequency of each label in the node set. The generation of the edge set \( \mathcal{E}' \) in the condensed graph depends on the similarity measure between nodes. For any two nodes \( v_i' \) and \( v_j' \), the presence of an edge between them depends on their similarity. If the similarity $s_{ij}$
 between the nodes exceeds a predefined threshold $\theta$, an edge is considered to exist between the nodes. The similarity $s_{ij}$ is computed by:
 \begin{equation}
 	s_{ij} = \text{sim}(v_i',  v_j') = \frac{v_i' \cdot  v_j'}{\|v_i'\| \| v_j'\|}
 \end{equation}
 Finally, the condensed graph can be represented as \( G' = (V', \mathcal{E}') \), where the edge set \( \mathcal{E}' \) is formed by the following rule:
 \begin{equation} \mathcal{E}' = \{ (v_i', v_j', s_{ij}) \mid s_{ij} \geq \theta \} \end{equation}
 Here, $\theta$ is the similarity threshold, used to ensure that only edges with higher similarity are retained. Our approach utilizes CGC for graph condensation. Unlike methods reliant on gradient matching or bi-level optimization, CGC introduces a training-free paradigm that transforms the objective into a class-to-node distribution matching problem. This is efficiently solved as a class partition task using clustering algorithms to generate condensed node features.

 \vspace{5pt}
 
 \noindent\textbf{Historical Graph Embeddings Generation} To replay information from the condensed historical graph to the model, we generate historical graph embeddings using a training approach with multiple condensed graphs. We feed a series of condensed graphs into a dynamic graph model and extract embeddings from the final condensed graph's features, with the methodology formalized as follows:
 
 Let $\left\{ G_C^{(1)}, G_C^{(2)}, \ldots, G_C^{(T)} \right\}$ represent the sequence of condensed historical graphs. Each condensed graph can be represented as $G_C^{(t)} = (V_C^{(t)}, \mathcal{E}_C^{(t)}, X_C^{(t)})$. Taking EvolveGCN\cite{b84} as an example, for the sequence 
$\left\{G_C^{(t)}, G_C^{(2)}, \ldots, G_C^{(T)} \right\}$, the model parameters $\theta$ are updated at each time step t as follows:
\begin{equation} \theta^{(t)} = \text{GRU}(\theta^{(t-1)}, \phi(G_C^{(t)})), \quad t = 1, 2, \ldots, T \end{equation}
where: $\theta^{(t)}$ represents the updated model parameters at time step t, $\theta^{(t-1)}$ denotes the parameters from the previous time step, $\phi(G_C^{(t)})$ is a feature extraction function that extracts features from the condensed graph $\phi(G_C^{(t)})$
, GRU is the Gated Recurrent Unit used to update parameters based on current graph features and previous parameters.Through this sequential training process, the model parameters 
$\theta$ gradually adapt to the evolutionary patterns in the condensed graph sequence. After training completion, the historical graph embeddings are generated using the final parameters $\theta^{(T)}$ and the last condensed graph, expressed as $f_{\theta^{(T)}}(V(G_C^{(T)}))$.
\vspace{5pt}

\noindent\textbf{Concatenation of Historical and Current Embeddings} To effectively integrate historical knowledge with current graph structural information, we propose an embedding fusion method based on feature dimension concatenation. Let $\mathbf{H}_{\text{current}} \in \mathbb{R}^{n \times d_n}$
represent the current node embeddings generated by the graph neural network, where 
n denotes the number of nodes and $d_n$ represents the node embedding dimension. The historical embeddings $\mathbf{H}_{\text{historical}} \in \mathbb{R}^{n \times d_h}$
are extracted from the final condensed graph using the trained dynamic graph model:
\begin{equation} \mathbf{H}_{\text{historical}} = f_{\theta^{(T)}}(V(G_C^{(T)})) \end{equation}
By performing concatenation along the feature dimension, the two embedding representations are fused into a unified feature representation:
\begin{equation} \mathbf{H}_{\text{combined}} = \text{Concat}(\mathbf{H}_{\text{current}}, \mathbf{H}_{\text{historical}}) \in \mathbb{R}^{n \times (d_n + d_h)} \end{equation}
To address the node matching problem between the current graph and the condensed historical graph, a cosine similarity-based threshold method is adopted: if the similarity between a current node and any node of condensed historical graph exceeds the threshold, their representations are concatenated; otherwise, zero-padding is applied. The resulting embeddings $\mathbf{H}_{\text{combined}}$ can be directly used for downstream tasks. This approach enables the model to simultaneously leverage structural patterns learned from historical graph data and specific features of the current graph.
\begin{table*}[t]
	\centering
	\setlength{\tabcolsep}{2pt}
	\scalebox{1}{%
		\begin{tabular}{c|cc|cc|cc|cc}
			\hline
			\multirow{2}{*}{\textbf{Method}} & \multicolumn{2}{c|}{\textbf{Arxiv}} & \multicolumn{2}{c|}{\textbf{Paper100M}} & \multicolumn{2}{c|}{\textbf{DBLP}} & \multicolumn{2}{c}{\textbf{Elliptic}} \\
			\cline{2-9}
			& PM & FM & PM & FM & PM & FM & PM & FM \\
			\hline
			TWP & $77.59\pm0.51\%$ & $4.07\pm0.10\%$ & $73.91\pm0.18\%$ & $6.84\pm0.04\%$ & $49.44\pm2.81\%$ & $7.06\pm1.57\%$ & $\mathbf{98.14\pm0.07\%}$ & $0.19\pm0.01\%$ \\
			ContinualGNN & $43.48\pm0.74\%$ & $31.48\pm0.10\%$ & $36.50\pm0.44\%$ & $28.92\pm2.54\%$ & $\mathbf{60.69\pm0.13\%}$ & $7.90\pm3.26\%$ & $93.12\pm0.30\%$ & $1.35\pm0.34\%$ \\
			CCC & $\mathbf{77.67\pm0.06\%}$ & $\mathbf{2.90\pm0.16\%}$ & $\mathbf{74.10\pm0.50\%}$ & $\mathbf{5.65\pm0.09\%}$ & $54.26\pm0.72\%$ & $\mathbf{4.90\pm0.27\%}$ & $97.74\pm0.07\%$ & $\mathbf{0.12\pm0.00\%}$ \\
			\hline
		\end{tabular}
	}
	\caption{Node Classification performance of CCC and baselines on Arxiv, Paper100M, DBLP and Elliptic.The higher PM means better performance whereas the lower the better for FM.}
	\label{tab:4x4}
\end{table*}

\begin{figure}[h]  
	\centering
	\includegraphics[width=0.6\linewidth]{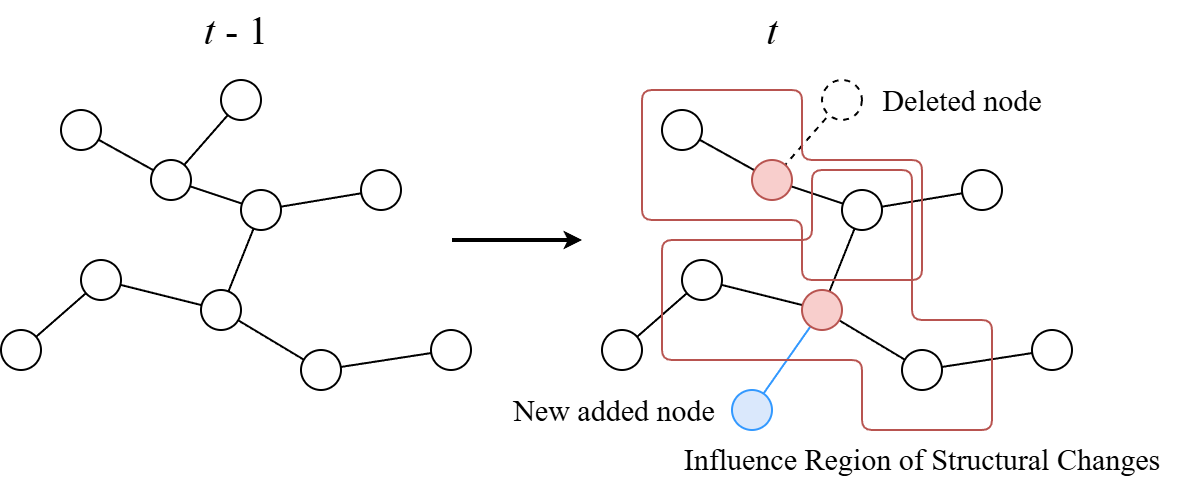}  
	\caption{Illustration of influence region of structural changes. The area enclosed by the red curve is the influence region of structural changes.}
	\label{fig:example}
\end{figure}

\noindent\textbf{Selective Historical Replay}
To address the issue that indiscriminate historical information replay can negatively impact nodes minimally affected by structural changes, we introduce a selective replay mechanism that confines historical embedding concatenation to significantly affected nodes. Our approach first detects structural changes by comparing the current and previous graphs to identify added/removed nodes and edges. The influence region is then defined as k-hop subgraphs centered around nodes directly connected to these topological modifications, as shown in Figure 2. Historical embeddings are selectively concatenated only for nodes within this identified change region, aiming for effective knowledge transfer while maintaining model accuracy.The formal definition is:
\begin{equation} \mathcal{R}_{\text{change}} = \bigcup_{v \in \mathcal{S}} \{u \in V_t \mid d(u,v) \leq k\} \end{equation}
where $\mathcal{S}$ denotes nodes adjacent to topological modifications and $d(u,v)$ represents the shortest path distance in $G_t$. Historical embeddings are exclusively concatenated for nodes within $\mathcal{R}_{\text{change}}$.

\section{Experiments}
\noindent In this section, we present the experimental results of CCC on four real-world datasets.
\subsection*{A. Setup}
\begin{enumerate}[label=\arabic*) , left=0pt, labelsep=0pt, wide=0pt, itemindent=0pt, listparindent=0pt]
\item \noindent Datasets.We conduct experiments on three real-world graph datasets with diverse application scenarios: DBLP, Arxiv, Elliptic and Paper100M. All four datasets are streaming graphs.
\item Baselines.The following baselines are compared:\\
• ContinualGNN\cite{b85} alleviates catastrophic forgetting by detecting new patterns via influence propagation and consolidating existing knowledge through multi-view replay and regularization.\\
• TWP mitigates catastrophic forgetting by preserving the topological structure of the graph and stabilizing important parameters.
\item Metrics.Two evaluation metrics are adopted: Performance Mean (PM) and Forgetting Measure (FM). PM measures the average model performance:
\begin{equation}PM = \frac{1}{T} \sum_{i=1}^{T} a_i\end{equation}
where \(T\) is the total number of tasks, and \(a_i\) refers to the accuracy of model on task \(i\).\\
\noindent \hspace*{1em}To adapt to dynamic graph scenarios, we modify the traditional forgetting metric. FM measures the ratio of nodes that were correctly predicted in the previous task but incorrectly predicted in the next task. It is defined as the proportion of nodes that are in both the set of correctly predicted nodes in task i and the set of incorrectly predicted nodes in task \textit{i}+1:
\begin{equation}FM = \frac{|C_i \cap E_{i+1}|}{|C_i|}\end{equation}
Where $C_i$ represents the set of nodes that were correctly predicted in task \textit{i}, $E_{i+1}$
represents the set of nodes that were incorrectly predicted in task \textit{i}+1 denotes the intersection of the two sets, i.e., the nodes that were correctly predicted in task i but incorrectly predicted in task \textit{i}+1.
\subsection*{B. Overall Results}
As summarized in Table \textrm{I}, the performance of CCC against baselines leads to the following observations.
\begin{itemize}[leftmargin=*, labelsep=0.5em]
	
\item Our proposed CCC method demonstrates consistently lower Forgetting Measure (FM) values than the baselines across all four datasets. This suggests that it effectively mitigates catastrophic forgetting.

\item The PM results further support CCC's strong overall performance. Its highly competitive accuracy on most benchmarks, with only minimal deviations from the top baseline, is complemented by its superior forgetting mitigation, as established by the FM metric.

\end{itemize}
\section{Discussion}
\setlength{\parindent}{1em}The improved performance of CCC in the Forgetting Measure (FM) can be attributed to its targeted strategy for mitigating catastrophic forgetting. Unlike baseline methods that adopt broad preservation strategies, CCC aims to identify nodes that are more susceptible to representation drift resulting from structural changes. By selectively replaying condensed historical information to these affected nodes, it helps reinforce and stabilize their representations, thus contributing to the alleviation of forgetting at its origin.
\section{Conclusion}
\setlength{\parindent}{1em}We propose CCC for dynamic graph continual learning, aiming to address node representation instability from structural changes. CCC detects topological changes in k-hop neighborhoods and selectively replays historical information from compressed graphs to affected nodes. Experiments on four datasets demonstrate superior performance.

\end{enumerate}


\begin{thebibliography}{00}
\bibitem{b1} F. Scarselli, M. Gori, A. C. Tsoi, M. Hagenbuchner, and G. Monfardini, 
``The graph neural network model,'' \textit{IEEE Transactions on Neural Networks}, 
vol. 20, no. 1, pp. 61--80, Jan. 2009.
\bibitem{b2} T. N. Kipf and M. Welling, 
``Semi-supervised classification with graph convolutional networks,'' 
in \textit{International Conference on Learning Representations (ICLR)}, 2017.
\bibitem{b3} W. L. Hamilton, R. Ying, and J. Leskovec, 
``Inductive representation learning on large graphs,'' 
in \textit{Advances in Neural Information Processing Systems (NeurIPS)}, 
pp. 1024--1034, 2017.
\bibitem{b4} L. Wang, K. Liu, and Y. Yuan, ``GT-A2T: Graph tensor alliance attention network,'' \textit{IEEE/CAA J. Autom. Sinica}, 2024, doi: 10.1109/JAS.2024.124863.
\bibitem{b5} M. Han, L. Wang, Y. Yuan, and X. Luo, 
``SGD-DyG: Self-Reliant Global Dependency Apprehending on Dynamic Graphs,'' 
\textit{Proc. ACM SIGKDD Conference on Knowledge Discovery and Data Mining}, 
pp. 1--10, 2025.
\bibitem{b6} J. Li, Y. Yuan, and X. Luo, 
``Learning Error Refinement in Stochastic Gradient Descent-based Latent Factor Analysis via Diversified PID Controllers,'' 
\textit{IEEE Transactions on Emerging Topics in Computational Intelligence}, 
vol. 9, no. 3, pp. 1--12, 2025.
\bibitem{b7} Y. Yuan, Y. Wang, and X. Luo, 
``A Node-Collaboration-Informed Graph Convolutional Network for Highly Accurate Representation to Undirected Weighted Graph,'' 
\textit{IEEE Transactions on Neural Networks and Learning Systems}, 
vol. 36, no. 6, pp. 11507--11519, 2025.
\bibitem{b8} Y. Yuan, S. Lu, and X. Luo, 
``A Proportional Integral Controller-Enhanced Non-negative Latent Factor Analysis Model,'' 
\textit{IEEE/CAA Journal of Automatica Sinica}, 
vol. 12, no. 6, pp. 1246--1259, 2024.
\bibitem{b9} Y. Yuan, J. Li, and X. Luo, 
``A Fuzzy PID-Incorporated Stochastic Gradient Descent Algorithm for Fast and Accurate Latent Factor Analysis,'' 
\textit{IEEE Transactions on Fuzzy Systems}, 
vol. 32, no. 7, pp. 4049--4061, 2024.
\bibitem{b10} J. Chen, Y. Yuan, and X. Luo, 
``SDGNN: Symmetry-Preserving Dual-Stream Graph Neural Networks,'' 
\textit{IEEE/CAA Journal of Automatica Sinica}, 
vol. 11, no. 7, pp. 1717--1719, 2024.

\bibitem{b11} X. Luo, J. Chen, Y. Yuan, and Z. Wang, 
``Pseudo Gradient-Adjusted Particle Swarm Optimization for Accurate Adaptive Latent Factor Analysis,'' 
\textit{IEEE Transactions on Systems Man Cybernetics: Systems}, 
vol. 54, no. 4, pp. 2213--2226, 2024.

\bibitem{b12} J. Li, X. Luo, Y. Yuan, and S. Gao, 
``A Nonlinear PID-Incorporated Adaptive Stochastic Gradient Descent Algorithm for Latent Factor Analysis,'' 
\textit{IEEE Transactions on Automation Science and Engineering}, 
vol. 21, no. 3, pp. 3742--3756, 2024.

\bibitem{b13} J. Chen, K. Liu, X. Luo, Y. Yuan, K. Sedraoui, Y. Al-Turki, and M. Zhou, 
``A State-migration Particle Swarm Optimizer for Adaptive Latent Factor Analysis of High-Dimensional and Incomplete Data,'' 
\textit{IEEE/CAA Journal of Automatica Sinica}, 
vol. 11, no. 11, pp. 2220--2235, 2024.

\bibitem{b14} Y. Yuan, X. Luo, M. Shang, and Z. Wang, 
``A Kalman-Filter-Incorporated Latent Factor Analysis Model for Temporally Dynamic Sparse Data,'' 
\textit{IEEE Transactions on Cybernetics}, 
vol. 53, no. 9, pp. 5788--5801, 2023.

\bibitem{b15} Y. Yuan, R. Wang, G. Yuan, and X. Luo, 
``An Adaptive Divergence-based Non-negative Latent Factor Model,'' 
\textit{IEEE Transactions on System Man Cybernetics: Systems}, 
vol. 53, no. 10, pp. 6475--6487, 2023.

\bibitem{b16} Y. Yuan, Q. He, X. Luo, and M. Shang, 
``A Multilayered-and-Randomized Latent Factor Model for High-Dimensional and Sparse Matrices,'' 
\textit{IEEE Transactions on Big Data}, 
vol. 8, no. 3, pp. 784--794, 2022.

\bibitem{b17} X. Luo, Y. Yuan, S. Chen, N. Zeng, and Z. Wang, 
``Position-Transitional Particle Swarm Optimization-Incorporated Latent Factor Analysis,'' 
\textit{IEEE Transactions on Knowledge and Data Engineering}, 
vol. 34, no. 8, pp. 3958--3970, 2022.

\bibitem{b18} M. Shang, Y. Yuan, X. Luo, and M. Zhou, 
``An $\alpha$-$\beta$-divergence-generalized Recommender for Highly-accurate Predictions of Missing User Preferences,'' 
\textit{IEEE Transactions on Cybernetics}, 
vol. 52, no. 8, pp. 8006--8018, 2022.

\bibitem{b19} X. Luo, Y. Yuan, M. Zhou, Z. Liu, and M. Shang, 
``Non-negative Latent Factor Model based on $\beta$-divergence for Recommender Systems,'' 
\textit{IEEE Transactions on System Man Cybernetics: Systems}, 
vol. 51, no. 8, pp. 4612--4623, 2021.

\bibitem{b20} Y. Yuan, M. Shang, and X. Luo, 
``Temporal Web Service QoS Prediction via Kalman Filter-Incorporated Dynamic Latent Factor Analysis,'' 
\textit{Proc. European Conference on Artificial Intelligence}, 
pp. 561--568, 2020.

\bibitem{b21} J. Chen, Y. Yuan, T. Ruan, J. Chen, and X. Luo, 
``Hyper-Parameter-Evolutionary Latent Factor Analysis for High-Dimensional and Sparse Data from Recommender Systems,'' 
\textit{Neurocomputing}, 
vol. 421, pp. 316--328, 2020.

\bibitem{b22} J. Li, Y. Yuan, T. Ruan, J. Chen, and X. Luo, 
``A Proportional-Integral-Derivative-Incorporated Stochastic Gradient Descent-Based Latent Factor Analysis Model,'' 
\textit{Neurocomputing}, 
vol. 427, pp. 29--39, 2020.

\bibitem{b23} Y. Yuan, X. Luo, and M. Shang, 
``Effects of Preprocessing and Training Biases in Latent Factor Models for Recommender Systems,'' 
\textit{Neurocomputing}, 
vol. 275, pp. 2019--2030, 2018.

\bibitem{b24} Z. He, M. Lin, X. Luo, and Z. Xu, 
``Structure-Preserved Self-Attention for Fusion Image Information in Multiple Color Spaces,'' 
\textit{IEEE Transactions on Neural Networks and Learning Systems}, 
2024, doi: 10.1109/TNNLS.2024.3490800.

\bibitem{b25} D. Wu, Y. Hu, K. Liu, J. Li, X. Wang, S. Deng, N. Zheng, and X. Luo, 
``An Outlier-Resilient Autoencoder for Representing High-Dimensional and Incomplete Data,'' 
\textit{IEEE Transactions on Emerging Topics in Computing}, 
2024, doi: 10.1109/TETCI.2024.3437370.

\bibitem{b26} D. Wu, Z. Li, F. Chen, J. He, and X. Luo, 
``Online Sparse Streaming Feature Selection With Gaussian Copula,'' 
\textit{IEEE Transactions on Big Data}, 
2024, doi: 10.1109/TBDATA.2024.XXXXXXX.

\bibitem{b27} M. Chen, L. Tao, J. Lou, and X. Luo, 
``Latent Factorization of Tensors Incorporated Battery Cycle Life Prediction,'' 
\textit{IEEE/CAA Journal of Automatica Sinica}, 
2024, doi: 10.1109/JAS.2024.124602.

\bibitem{b28} D. Wu, Z. Li, Z. Yu, Y. He, and X. Luo, 
``Robust Low-rank Latent Feature Analysis for Spatio-Temporal Signal Recovery,'' 
\textit{IEEE Transactions on Neural Networks and Learning Systems}, 
2023, doi: 10.1109/TNNLS.2023.3339786.

\bibitem{b29} P. Tang and X. Luo, 
``Neural Tucker Factorization,'' 
\textit{IEEE/CAA Journal of Automatica Sinica}, 
vol. 12, no. 2, pp. 475--477, 2025.

\bibitem{b30} M. Lin, J. Liu, H. Chen, X. Xu, X. Luo, and Z. Xu, 
``A 3D Convolution-Incorporated Dimension Preserved Decomposition Model for Traffic Data Prediction,'' 
\textit{IEEE Transactions on Intelligent Transportation Systems}, 
vol. 26, no. 1, pp. 673--690, 2025.

\bibitem{b31} H. Yang, M. Lin, H. Chen, X. Luo, and Z. Xu, 
``Latent Factor Analysis Model with Temporal Regularized Constraint for Road Traffic Data Imputation,'' 
\textit{IEEE Transactions on Intelligent Transportation Systems}, 
vol. 26, no. 1, pp. 724--741, 2025.

\bibitem{b32} X. Liao, K. Hoang, and X. Luo, 
``Local Search-based Anytime Algorithms for Continuous Distributed Constraint Optimization Problems,'' 
\textit{IEEE/CAA Journal of Automatica Sinica}, 
vol. 12, no. 1, pp. 1--3, 2024.

\bibitem{b33} H. Wu, Y. Qiao, and X. Luo, 
``A Fine-Grained Regularization Scheme for Nonnegative Latent Factorization of High-Dimensional and Incomplete Tensors,'' 
\textit{IEEE Transactions on Services Computing}, 
vol. 17, no. 6, pp. 3006--3021, 2024.

\bibitem{b34} T. Chen, W. Yang, Z. Zhang, and X. Luo, 
``An Efficient Industrial Robot Calibrator with Multi-Planer Constraints,'' 
\textit{IEEE Transactions on Industrial Informatics}, 
vol. 20, no. 12, pp. 14341--14350, 2024.

\bibitem{b35} W. Yang, S. Li, and X. Luo, 
``Data Driven Vibration Control: A Review,'' 
\textit{IEEE/CAA Journal of Automatica Sinica}, 
vol. 11, no. 9, pp. 1898--1917, 2024.

\bibitem{b36} Y. Zhong, K. Liu, S. Gao, and X. Luo, 
``Alternating-Direction-Method of Multipliers-based Adaptive Nonnegative Latent Factor Analysis,'' 
\textit{IEEE Transactions on Emerging Topics in Computing}, 
vol. 8, no. 5, pp. 3544--3558, 2024.

\bibitem{b37} T. Chen, S. Li, Y. Qiao, and X. Luo, 
``A Robust and Efficient Ensemble of Diversified Evolutionary Computing Algorithms for Accurate Robot Calibration,'' 
\textit{IEEE Transactions on Instrumentation and Measurement}, 
vol. 73, pp. 1--14, 2024.

\bibitem{b38} N. Zeng, X. Li, P. Wu, H. Li, and X. Luo, 
``A Novel Tensor Decomposition-based Efficient Detector for Low-altitude Aerial Objects with Knowledge Distillation Scheme,'' 
\textit{IEEE/CAA Journal of Automatica Sinica}, 
vol. 11, no. 2, pp. 487--501, 2024.

\bibitem{b39} Z. Li, S. Li, and X. Luo, 
``A Novel Machine Learning System for Industrial Robot Arm Calibration,'' 
\textit{IEEE Transactions on Circuits and Systems II: Express Briefs}, 
vol. 71, no. 4, pp. 2364--2368, 2024.

\bibitem{b40} D. Wu, P. Zhang, Y. He, and X. Luo, 
``MMLF: Multi-Metric Latent Feature Analysis for High-Dimensional and Incomplete Data,'' 
\textit{IEEE Transactions on Services Computing}, 
vol. 17, no. 2, pp. 575--588, 2024.

\bibitem{b41} W. Qin, X. Luo, and M. Zhou, 
``Adaptively-accelerated Parallel Stochastic Gradient Descent for High-Dimensional and Incomplete Data Representation Learning,'' 
\textit{IEEE Transactions on Big Data}, 
vol. 10, no. 1, pp. 92--107, 2024.

\bibitem{b42} W. Li, R. Wang, and X. Luo, 
``A Generalized Nesterov-Accelerated Second-Order Latent Factor Model for High-Dimensional and Incomplete Data,'' 
\textit{IEEE Transactions on Neural Networks and Learning Systems}, 
vol. 36, no. 1, pp. 1518--1532, 2024.

\bibitem{b43} F. Bi, T. He, and X. Luo, 
``A Fast Nonnegative Autoencoder-based Approach to Latent Feature Analysis on High-Dimensional and Incomplete Data,'' 
\textit{IEEE Transactions on Services Computing}, 
vol. 17, no. 3, pp. 733--746, 2024.

\bibitem{b44} W. Qin and X. Luo, 
``Asynchronous Parallel Fuzzy Stochastic Gradient Descent for High-Dimensional Incomplete Data,'' 
\textit{IEEE Transactions on Fuzzy Systems}, 
vol. 32, no. 2, pp. 445--459, 2024.

\bibitem{b45} Z. Liu, X. Luo, and M. Zhou, 
``Symmetry and Graph Bi-regularized Non-Negative Matrix Factorization for Precise Community Detection,'' 
\textit{IEEE Transactions on Automation Science and Engineering}, 
vol. 21, no. 2, pp. 1406--1420, 2024.

\bibitem{b46} D. Wu, X. Luo, Y. He, and M. Zhou, 
``A Prediction-sampling-based Multilayer-structured Latent Factor Model for Accurate Representation to High-dimensional and Sparse Data,'' 
\textit{IEEE Transactions on Neural Networks and Learning Systems}, 
vol. 35, no. 3, pp. 3845--3858, 2024.

\bibitem{b47} M. Chen, Y. Qiao, R. Wang, and X. Luo, 
``A Generalized Nesterov's Accelerated Gradient-Incorporated Non-negative Latent-factorization-of-tensors Model for Efficient Representation to Dynamic QoS Data,'' 
\textit{IEEE Transactions on Emerging Topics in Computational Intelligence}, 
vol. 8, no. 3, pp. 2386--2400, 2024.

\bibitem{b48} L. Jin, Y. Li, X. Zhang, and X. Luo, 
``Fuzzy k-Winner-Take-All Network for Competitive Coordination in Multi-robot Systems,'' 
\textit{IEEE Transactions on Fuzzy Systems}, 
vol. 32, no. 4, pp. 2005--2016, 2024.

\bibitem{b49} Z. Xie, L. Jin, X. Luo, M. Zhou, and Y. Zheng, 
``A Biobjective Scheme for Kinematic Control of Mobile Robotic Arms with Manipulability Optimization,'' 
\textit{IEEE/ASME Transactions on Mechatronics}, 
vol. 29, no. 2, pp. 1534--1545, 2024.

\bibitem{b50} W. Qin, X. Luo, S. Li, and M. Zhou, 
``Parallel Adaptive Stochastic Gradient Descent Algorithms for Latent Factor Analysis of High-Dimensional and Incomplete Industrial Data,'' 
\textit{IEEE Transactions on Automation Science and Engineering}, 
vol. 21, no. 3, pp. 2716--2729, 2024.

\bibitem{b51} J. Yan, L. Jin, X. Luo, and S. Li, 
``Modified RNN for Solving Comprehensive Sylvester Equation with TDOA Application,'' 
\textit{IEEE Transactions on Neural Networks and Learning Systems}, 
vol. 35, no. 9, pp. 12553--12563, 2024.

\bibitem{b52} L. Wei, L. Jin, and X. Luo, 
``A Robust Coevolutionary Neural-Based Optimization Algorithm for Constrained Nonconvex Optimization,'' 
\textit{IEEE Transactions on Neural Networks and Learning Systems}, 
vol. 35, no. 6, pp. 7778--7791, 2024.

\bibitem{b53} Q. Jiang, D. Liu, H. Zhu, S. Wu, N. Wu, X. Luo, and Y. Qiao, 
``Iterative Role Negotiation via the Bi-level GRA++ with Decision Tolerance,'' 
\textit{IEEE Transactions on Computational Social Systems}, 
vol. 11, no. 6, pp. 7484--7499, 2024.

\bibitem{b54} J. Wang, W. Li, and X. Luo, 
``A Distributed Adaptive Second-order Latent Factor Analysis Model,'' 
\textit{IEEE/CAA Journal of Automatica Sinica}, 
vol. 11, no. 11, pp. 2343--2345, 2024.

\bibitem{b55} J. Li, F. Tan, C. He, Z. Wang, H. Song, P. Hu, and X. Luo, 
``Saliency-Aware Dual Embedded Attention Network for Multivariate Time-Series Forecasting in Information Technology Operations,'' 
\textit{IEEE Transactions on Industrial Informatics}, 
vol. 20, no. 3, pp. 4206--4217, 2024.

\bibitem{b56} X. Luo, L. Wang, P. Hu, and L. Hu, 
``Predicting Protein-Protein Interactions Using Sequence and Network Information via Variational Graph Autoencoder,'' 
\textit{IEEE/ACM Transactions on Computational Biology and Bioinformatics}, 
vol. 20, no. 5, pp. 3182--3194, 2023.

\bibitem{b57} X. Luo, H. Wu, and Z. Li, 
``NeuLFT: A Novel Approach to Nonlinear Canonical Polyadic Decomposition on High-Dimensional Incomplete Tensors,'' 
\textit{IEEE Transactions on Knowledge and Data Engineering}, 
vol. 35, no. 6, pp. 6148--6166, 2023.

\bibitem{b58} X. Luo, Y. Zhong, Z. Wang, and M. Li, 
``An Alternating-direction-method of Multipliers-Incorporated Approach to Symmetric Non-negative Latent Factor Analysis,'' 
\textit{IEEE Transactions on Neural Networks and Learning Systems}, 
vol. 34, no. 8, pp. 4826--4840, 2023.

\bibitem{b59} X. Luo, Y. Zhou, Z. Liu, and M. Zhou, 
``Fast and Accurate Non-negative Latent Factor Analysis on High-dimensional and Sparse Matrices in Recommender Systems,'' 
\textit{IEEE Transactions on Knowledge and Data Engineering}, 
vol. 35, no. 4, pp. 3897--3911, 2023.

\bibitem{b60} D. Wu, Y. He, and X. Luo, 
``A Graph-incorporated Latent Factor Analysis Model for High-dimensional and Sparse Data,'' 
\textit{IEEE Transactions on Emerging Topics in Computing}, 
vol. 11, no. 4, pp. 907--917, 2023.

\bibitem{b61} L. Chen and X. Luo, 
``Tensor Distribution Regression based on the 3D Conventional Neural Networks,'' 
\textit{IEEE/CAA Journal of Automatica Sinica}, 
vol. 10, no. 7, pp. 1628--1630, 2023.

\bibitem{b62} F. Bi, X. Luo, B. Shen, H. Dong, and Z. Wang, 
``Proximal Alternating-Direction-Method-of-Multipliers-Incorporated Nonnegative Latent Factor Analysis,'' 
\textit{IEEE/CAA Journal of Automatica Sinica}, 
vol. 10, no. 6, pp. 1388--1406, 2023.

\bibitem{b63} L. Hu, Y. Yang, Z. Tang, Y. He, and X. Luo, 
``FCAN-MOPSO: An Improved Fuzzy-based Graph Clustering Algorithm for Complex Networks with Multi-objective Particle Swarm Optimization,'' 
\textit{IEEE Transactions on Fuzzy Systems}, 
vol. 31, no. 10, pp. 3470--3484, 2023.

\bibitem{b64} F. Bi, T. He, Y. Xie, and X. Luo, 
``Two-Stream Graph Convolutional Network-Incorporated Latent Feature Analysis,'' 
\textit{IEEE Transactions on Services Computing}, 
vol. 16, no. 4, pp. 3027--3042, 2023.

\bibitem{b65} W. Yang, S. Li, Z. Li, and X. Luo, 
``Highly-Accurate Manipulator Calibration via Extended Kalman Filter-Incorporated Residual Neural Network,'' 
\textit{IEEE Transactions on Industrial Informatics}, 
vol. 19, no. 11, pp. 10831--10841, 2023.

\bibitem{b66} Z. Liu, Y. Yi, and X. Luo, 
``A High-Order Proximity-Incorporated Nonnegative Matrix Factorization-based Community Detector,'' 
\textit{IEEE Transactions on Emerging Topics in Computational Intelligence}, 
vol. 7, no. 3, pp. 700--714, 2023.

\bibitem{b67} W. Li, X. Luo, H. Yuan, and M. Zhou, 
``A Momentum-accelerated Hessian-vector-based Latent Factor Analysis Model,'' 
\textit{IEEE Transactions on Services Computing}, 
vol. 16, no. 2, pp. 830--844, 2023.

\bibitem{b68} D. Wu, P. Zhang, Y. He, and X. Luo, 
``A Double-Space and Double-Norm Ensembled Latent Factor Model for Highly Accurate Web Service QoS Prediction,'' 
\textit{IEEE Transactions on Services Computing}, 
vol. 16, no. 2, pp. 802--814, 2023.

\bibitem{b69} L. Jin, S. Liang, X. Luo, and M. Zhou, 
``Distributed and Time-Delayed k-Winner-Take-All Network for Competitive Coordination of Multiple Robots,'' 
\textit{IEEE Transactions on Cybernetics}, 
vol. 53, no. 1, pp. 641--652, 2023.

\bibitem{b70} Z. Li, S. Li, O. Bamasag, A. Alhothali, and X. Luo, 
``Diversified Regularization Enhanced Training for Effective Manipulator Calibration,'' 
\textit{IEEE Transactions on Neural Networks and Learning Systems}, 
vol. 34, no. 11, pp. 8778--8790, 2023.

\bibitem{b71} X. Xu, M. Lin, X. Luo, and Z. Xu, 
``HRST-LR: A Hessian Regularization Spatio-Temporal Low Rank Algorithm for Traffic Data Imputation,'' 
\textit{IEEE Transactions on Intelligent Transportation Systems}, 
vol. 24, no. 10, pp. 11001--11017, 2023.

\bibitem{b72} Y. Zhou, X. Luo, and M. Zhou, 
``Cryptocurrency Transaction Network Embedding from Static and Dynamic Perspectives: An Overview,'' 
\textit{IEEE/CAA Journal of Automatica Sinica}, 
vol. 10, no. 5, pp. 1105--1121, 2023.

\bibitem{b73} Z. Li, X. Luo, and S. Li, 
``Efficient Industrial Robot Calibration via a Novel Unscented Kalman Filter-Incorporated Variable Step-Size Levenberg-Marquardt Algorithm,'' 
\textit{IEEE Transactions on Instrumentation and Measurement}, 
vol. 72, pp. 1--12, 2023.

\bibitem{b74} M. Chen, C. He, and X. Luo, 
``MNL: A Highly-Efficient model for Large-scale Dynamic Weighted Directed Network Representation,'' 
\textit{IEEE Transactions on Big Data}, 
vol. 9, no. 3, pp. 889--903, 2023.

\bibitem{b75} W. Li, R. Wang, X. Luo, and M. Zhou, 
``A Second-order Symmetric Non-negative Latent Factor Model for Undirected Weighted Network Representation,'' 
\textit{IEEE Transactions on Network Science and Engineering}, 
vol. 10, no. 2, pp. 606--618, 2023.

\bibitem{b76} Z. Xie, L. Jin, and X. Luo, 
``Kinematics-Based Motion-Force Control for Redundant Manipulators with Quaternion Control,'' 
\textit{IEEE Transactions on Automation Science and Engineering}, 
vol. 20, no. 3, pp. 1815--1828, 2023.

\bibitem{b77} F. Zhang, L. Jin, and X. Luo, 
``Error-Summation Enhanced Newton Algorithm for Model Predictive Control of Redundant Manipulators,'' 
\textit{IEEE Transactions on Industrial Electronics}, 
vol. 70, no. 3, pp. 2800--2811, 2023.

\bibitem{b78} X. Chen, X. Luo, L. Jin, S. Li, and M. Liu, 
``Growing Echo State Network With An Inverse-Free Weight Update Strategy,'' 
\textit{IEEE Transactions on Cybernetics}, 
vol. 53, no. 2, pp. 753--764, 2023.

\bibitem{b79} J. Chen, R. Wang, D. Wu, and X. Luo, 
``A Differential Evolution-Enhanced Position-Transitional Approach to Latent Factor Analysis,'' 
\textit{IEEE Transactions on Emerging Topics in Computational Intelligence}, 
vol. 7, no. 2, pp. 389--401, 2023.

\bibitem{b80} X. Luo, Y. Zhou, Z. Liu, L. Hu, and M. Zhou, 
``Generalized Nesterov's Acceleration-incorporated, Non-negative and Adaptive Latent Factor Analysis,'' 
\textit{IEEE Transactions on Services Computing}, 
vol. 15, no. 5, pp. 2809--2823, 2022.
\bibitem{b81} H. Liu, Y. Yang, and X. Wang, 
``Overcoming catastrophic forgetting in graph neural networks,'' in \textit{Proceedings of the AAAI Conference on Artificial Intelligence}, 2021.
pp. 3158--3164, 2022.
\bibitem{b82} F. Zhou and C. Cao, ``Overcoming catastrophic forgetting in graph neural networks with experience replay,'' in \textit{Proceedings of the AAAI Conference on Artificial Intelligence}, vol. 34, pp. 1--8, April 2020.
\bibitem{b83} S. Choi, W. Kim, S. Kim, Y. In, S. Kim, and C. Park, ``DSLR: Diversity enhancement and structure learning for rehearsal-based graph continual learning,'' in \textit{Proceedings of the ACM Web Conference 2024}, Singapore, Singapore, pp. 1--12, May 2024.
\bibitem{b84} A. Pareja et al., 
``EvolveGCN: Evolving graph convolutional networks for dynamic graphs,'' 
in \textit{Proceedings of the AAAI Conference on Artificial Intelligence}, 
pp. 5363--5370, 2020.
\bibitem{b85} J. Wang, G. Song, Y. Wu, and L. Wang, 
``Streaming Graph Neural Networks via Continual Learning,'' 
in \textit{Proceedings of the 29th ACM International Conference on Information and Knowledge Management}, 
2020, pp. 1--10.
\end{thebibliography}
\end{document}